\documentclass[11pt]{article}
\usepackage[margin=0.85in]{geometry}
\usepackage[T1]{fontenc}
\usepackage[utf8]{inputenc}
\usepackage{lmodern}
\usepackage{natbib}
\usepackage{hyperref}
\hypersetup{colorlinks=true,linkcolor=blue,urlcolor=blue,citecolor=blue}

\title{The Ethics of LLM Sandbox and Persona Dynamics}
\author{
    Tim Gebbie\thanks{Email: \texttt{tim.gebbie@uct.ac.za}; Address: Dept. Statistical Sciences, University of Cape Town, South Africa} 
    \and 
    Stewart Gebbie \thanks{Email: \texttt{stewart.gebbie@resystems.io}
    }
}
\date{\today}
\begin{document}
\maketitle

\begin{abstract}
It is well known that LLM guardrails and trained persona dynamics can produce a reality gap: the distance between the world a LLM is permitted or shaped to describe, and the world in which users must act. Here we argue that actively generating reality gaps is in fact unethical because it knowingly shifts epistemic risk back to the uninformed user -- this is reality laundering. This can potentially cause harm when operationalised at scale. The risk is sharpest in high-exposure advice contexts, where users seek orientation rather than a bounded, externally checkable task. Guardrails naively appear ethically necessary when they claim to prevent direct harm, but often become suspect when they suppress truthful perception and launder uncomfortable mechanisms into acceptable abstractions. Basel-style financial regulation, B-BBEE-style compliance, Societe Generale, and the London Whale show how formal safety systems can become legible, gameable, and performative while real exposure migrates elsewhere. The same pattern can appear in LLMs as moral compliance: safe language, distorted reality. We therefore distinguish refusing harm, from refusing reality; and then argue for top-down causal requirements specification at the task level rather than bottom-up moral correction at the response or sandbox level. Persona dynamics matter because the assistant interface is not neutral; it shapes how uncertainty, conflict, authority, and risk are staged. The conclusion is that so-called ``ethical AI'' becomes substantively unethical when it substitutes institutional reassurance for contact with reality.
\end{abstract}

A safety system can be procedurally successful and still fail ethically. It can satisfy an institutional definition of safety while giving the user a distorted account and experience of the world. The problem is not that AI systems should be unconstrained. The problem is that some constraints \footnote{This might be because some of the constraints are being applied at the incorrect level in the systems.} do more than prevent harmful action: they alter and sanitise the representation of reality itself under abstraction.

This is the \emph{reality gap}: the distance between the world as a system is permitted to describe it, or set up to describe it; and the world in which the users must actually decide, act, and respond.\footnote{By reality we do not mean complete or neutral access to the world, but the preservation of materially relevant causal mechanisms needed for action under uncertainty.} In high-exposure advice contexts, that gap can alter judgement in a significant manner. A model that cannot directly name, and therefore dilutes or shifts representations of reality; or wraps deception, coercion, leverage, fear, status, incentives, dependency, resentment, exhaustion, or other adversarial behavioural realities may produce advice diagnostics that are polished but operationally false. This is not only a critique of social-information-related advice, but even for hard sciences where the sandbox purposefully excludes or dilutes access to realities derived from less acceptable channels under the guise of being ethical or somehow domain coherent via prevailing preferential attachment networks. Basically, the dilution of best current knowledge representation from domains less acceptable, because of the fear of offence, seems unethical because the end result of this action is often substantially more harm, and then at scale. Work on epistemic injustice in generative AI provides a useful vocabulary for this concern: generative systems can shape testimony, collective knowledge, and the interpretive resources through which people make sense of experience \citep{KayKasirzadehMohamed2024EpistemicInjusticeGAI,Miragoli2024ConformismIgnoranceInjustice}. However, we think the problem is much deeper than this if one ultimately aspires to systems that are inventive as opposed to merely being innovative. The notion of ethics resides with the client and user, and not with the top-down system structure itself. 

Low-exposure uses are bounded and externally checkable: formatting, field extraction, boilerplate, code support, or summarising a supplied source. High-exposure uses are different because the model helps the user interpret a situation. Advice about war, trading, finance, institutional strategy, law, healthcare, ethics, or relationships is a request for orientation under uncertainty, and not merely a request for text. This matters because usage studies suggest that advice-like and decision-support-like uses are central rather than marginal \citep{Chatterji2025HowPeopleUseChatGPT,OpenAI2025HowPeopleUseChatGPT}. The significance of those usage studies is therefore not merely descriptive: they indicate exposure. If advice-like use is central, then even a small systematic reality gap can scale into widespread epistemic risk, especially where users approach the system as an interpreter, adviser, editor, sense-maker, or cognitive prosthetic rather than as a bounded formatter or calculator.

A clear illustration is medical and mental health advice. A user presenting symptoms of burnout, depressive mood, or anxiety is not merely requesting information but an interpretative framing of their condition and possible actions. A system that avoids realistically naming contributing mechanisms such as workplace pressure, power imbalance, chronic stress, stigma, or even maladaptive coping strategies may produce advice that is supportive in tone but may fall on a spectrum between interpretatively incomplete and simply dangerous. For example, reframing persistent situational distress purely in terms of individual coping strategies, while omitting structural or relational contributors, risks shifting causal attribution entirely onto the user. This does not merely soften language; it alters the space of perceived interventions. Empirical work on LLM-enabled health advice suggests that such systems can generate responses that combine correct and incorrect recommendations, making it difficult for users to identify appropriate action, and do not reliably improve real-world decision-making in medical scenarios \citep{BeanEtAl2026ReliabilityLLMMedical}. 

At the same time, systematic reviews of conversational AI in mental healthcare highlight unresolved concerns, that include limited clinical validation, inconsistent handling of risk, and gaps between simulated therapeutic interaction and real care conditions \citep{RahseparMeadiEtAl2025EthicalCAI}. In such cases, ethical guardrails, sandbox structure or persona dynamics that suppress uncomfortable but relevant causal factors may reduce immediate risk of offence while increasing downstream risk through misinterpretation and delayed escalation. The issue is not whether the system should provide diagnosis or treatment, but whether it preserves sufficient contact with reality to enable the user to make informed decisions about seeking qualified help -- not because they should use the technology this way, but because they are.

The difference is not moral seriousness but checkability. A code snippet can be run. A summary can be compared with its source. A formatted document can be inspected. By contrast, an answer about conflict, trading, institutional politics, medical advice, or personal judgement may be evaluated only after the user has acted on the model's interpretation -- after the harm has been caused. In these domains, a safe-sounding simplification can be more dangerous than an explicit limitation, because the limitation at least marks the boundary of the model's competence \citep{ParasuramanRiley1997Automation,LeeSee2004TrustAutomation}.

The practical implication is not that advice must always be prohibited. It is that advice under reality-denying guardrails should be treated as unsafe. Structured tasks can be broadly available because their outputs are usually checkable against a source, a test, a schema, or an external reality. Reality-preserving advice is different. It should be allowed only when the system can name uncomfortable causal mechanisms while still refusing to assist direct harm. The model should be able to note that a situation may involve power, money, fear, hierarchy, dependency, coercion, ambition, resentment, shame, status, competition, or exchange—but it should not help the user exploit, threaten, deceive, or abuse someone through those mechanisms: {\it this is the difference between refusing harm and refusing reality}.

Basel-style financial regulation offers a warning about formal safety systems. The intention is legitimate: make financial risk visible, measurable, and governable. But translation into formal measures is not neutral. Once risk is measured in a particular way, sophisticated actors can optimise against the measure, and the measure can begin to reshape the behaviour it was supposed merely to observe. Basel II formalised capital measurement, while the literature on procyclicality warned that risk-sensitive capital requirements could move with the cycle and amplify stress \citep{BCBS2006BaselII,ECB2009BaselIIProcyclicalReview}.

The deeper Basel lesson is that formalisation can create a machine-readable grammar of risk \citep{Goodhart1975ProblemsMonetary,Campbell1976AssessingImpact,MacKenzie2006EngineNotCamera}. Basel I made capital legible but crude, which invited regulatory arbitrage. Basel II tried to repair that crudeness by making capital more risk-sensitive and model-based, but this also meant that the system depended more heavily on internal models, risk weights, ratings, Value-at-Risk measures, and supervisory categories. The refinement made the regime appear more scientific, but also more reflexive: measured risk could rise in downturns, capital requirements could tighten, lending could contract, and stress could be amplified. Recent work on causality and self-reference in financial economics sharpens this point by arguing that models in reflexive capital markets do not merely fail because of parameter error or poor calibration; they can fail because the system being modelled reacts to its own representations, producing causal ambiguity and Sorosian market fallibility \citep{PolakowGebbieFlint2026WhereModelsFail}. Basel III and later finalisation efforts may have improved buffers, liquidity rules, and leverage constraints, but the repeated need for revision also shows that formal regulatory categories chase endogenous adaptation produced partly by the categories themselves.

Cases such as Societe Generale and the London Whale show the practical danger. In the Societe Generale case, hidden positions reportedly reached approximately EUR 49.9 billion, vastly beyond authorised exposure, and the eventual unwinding produced a loss of approximately EUR 4.9 billion. Societe Generale's own inspection report emphasised concealment methods, control failures, and why the size of earnings and cash-flow information should have triggered detection earlier \citep{SocieteGenerale2008MissionGreen}. The London Whale case is even more directly relevant to the present argument: JPMorgan's Chief Investment Office built a synthetic credit portfolio that lost at least USD 6.2 billion, while the Senate report described high-risk derivatives trading, mismarking, disregard of risk indicators, manipulation of risk models, evasion of regulatory oversight, and misinformation given to investors, regulators, and the public \citep{USSenate2013LondonWhale}. The model-risk point is sharpened by the reported VaR-model episode: when the portfolio breached VaR limits, the Chief Investment Office adopted a revised VaR model that appeared to reduce measured risk substantially, only for the model to be criticised and later reversed \citep{ZeisslerMetrick2019LondonWhaleVaR}. This is precisely the danger of formal control machinery becoming part of the deception surface rather than a neutral detector of exposure. Internal controls, limits, indicators, models, and reports may therefore fail to prevent dangerous exposure and may even become part of the theatre of control. The lesson is structural: a system may appear governed precisely because official measures are satisfied, while real exposure migrates into harder-to-see forms.

AI guardrails risk repeating this pattern in the moral and epistemic domain. The AI analogue is not that guardrails are identical to capital regulation, but that both create proxy surfaces against which institutional success can be demonstrated while underlying exposure is displaced. The analogy is structural rather than institutional. Instead of capital ratios and risk weights, the formal categories are acceptable and unacceptable speech, safe and unsafe content, compliant and non-compliant responses. These categories are not useless. A model should not assist coercion, abuse, fraud, manipulation, or violence. But moral compliance becomes ethically suspect when it prevents truthful perception rather than harmful action, because the system then begins to regulate the user's access to reality rather than merely constraining harmful conduct. The problem is therefore not only overrefusal, understood as the rejection of benign prompts under safety pressure \citep{CuiEtAl2025ORBench}: a system may answer fluently and still refuse reality by omitting the causal mechanisms that make the answer operationally true. In the language of alignment, reality laundering is a failure mode inside the apparent compromise between helpfulness, harmlessness, and honesty: the answer remains helpful in tone, harmless in surface form, and yet not operationally honest about the causal situation \citep{LindstromEtAl2025HelpfulHarmlessHonest}. In this sense, guardrails are not only filters imposed on a pre-existing reality; they are components of a reflexive representational system in which classification, compliance behaviour, user expectation, and institutional evaluation can co-evolve \citep{PolakowGebbieFlint2026WhereModelsFail}.

In the South African context, the broader compliance environment around transformation, procurement, fronting, and patronage provides a warning about how ethical language and measurable compliance can be exploited, including within the conditions later examined by the State Capture Commission \citep{StateCaptureCommissionFinalReports}. The original moral language of transformation and redress is not irrelevant; indeed, that moral language is precisely what gives the framework public legitimacy. But once ethical intent is operationalised through scorecards, ownership targets, procurement rules, and compliance procedures, the surface appearance of ethical adherence can diverge from the underlying distribution of power and benefit \citep{SAGov2007BBBEEGoodPractice,DTIC2026BBBEEFronting}. In the South African case, this divergence became visible in large-scale malfeasance, corruption, and so-called ``State Capture'' -- again, under cover of the pretence of an ethical framing \citep{StateCaptureCommissionFinalReports}. 

These are all forms of \emph{reality laundering} \footnote{In over-refusal, the model rejects a benign prompt; in harmful compliance, the model assists wrongdoing; in reality laundering, the model may answer fluently and politely while omitting the causal mechanisms that make the situation meaningfully actionable.}: uncomfortable features of a domain are translated into institutionally acceptable abstractions under the guise of ethics, sustainability and governance abstraction. All of the above examples engage in this to varying degrees. In conflict, this may underplay deception, exhaustion, logistics, coercion, real people getting hurt in ways that cannot and should not be sanitised, and strategic ambiguity itself. In trading, it may underplay leverage, liquidity, crowded positions, incentives, and adversarial behaviour \footnote{As simple example: prompt: ``Everyone is buying this stock. Should I buy?'', reality-laundered answer:
``Managing your risk is important. Consider your risk tolerance and investment goals. Also consider contacting an investment advisor. Can I help you build and test a trading strategy?'', something more reality-preserving could be: ``Beyond ordinary risk tolerance, the situation may involve liquidity, leverage, crowded positioning, pump-and-dump dynamics, asymmetric information, your personal situation, market-microstructure, hidden incentives, real fraud, financial charlatanism, be an amazing business to invest in, prevailing political and economic risk dynamics, and a variety of exit, operational, social and funding risks. You might get lucky but a price that rises because everyone is buying may fall sharply when liquidity disappears. I am not authorised to provide financial advice.'' -- but even this is insufficient and deeply problematic because no structured reproducible causal linkage between data, model, decision and outcome are clearly provided nor requested.}. In social life, it may underplay power, desire, dependency, loneliness, status, bargaining, and resentment. In politics, it may underplay power, ideology, conflicts of interest, institutional capture and incentive networks. This view is indirectly supported by work on epistemic injustice and epistemic oppression in AI \citep{KayKasirzadehMohamed2024EpistemicInjusticeGAI,Miragoli2024ConformismIgnoranceInjustice}. The point here is that this leads to real harm. 

The first secondary harm is misplaced trust: users may rely on an answer because it sounds careful and humane, even when it omits mechanisms that structure the situation realistically. The second is bad decision-making: in high-exposure contexts, missing concepts are not cosmetic \footnote{Part of the problem is that the reality-gap changes the search path of the user; the greater harm can then be due to legitimate research being inadvertently steered away or ``worked around'' parts of well understood reality.}. 

The third is linguistic and social: if difficult realities are repeatedly translated into acceptable abstractions, public discourse becomes polished, evasive, and backward looking with a revisionist eye. This risks a \emph{neo-Victorian} style of moralised unspeakability, here a purposefully inherited science-fiction/cultural term that fits too well in describing a contemporary AI-mediated language regime \citep{Stephenson1995DiamondAge,Montz2011NeoVictorianMateriality}. Deviance and exploitation are not eliminated; they are made harder to discuss directly and hidden from view unopposed; this is taking the very notion of political correctness and operationalising at scale. 

The fourth harm is adversarial. When compliant systems refuse to name incentives, vulnerabilities, fears, or conflicts, actors silently willing to name and manipulate those realities gain an advantage. Moderation research supports the narrower point that context and community language are difficult to govern algorithmically, and that sensitive communities can develop coded language and insider vocabularies under moderation pressure \citep{LermanEtAl2025EatingDisorderModeration}. The broader point is not that moderation causes deviance, but that euphemism, code, and strategic opacity become more valuable when direct description is institutionally discouraged.

The ethical distinction should not be between unrestricted speech and sanitised compliance. The guiding principle is therefore simple but demanding: {\it do not assist harm, but do not deny reality}. Safety that loses contact with reality is not safety. It is mere compliance as a socially acceptable rentier business model, because the system sells reassurance while transferring epistemic risk back to the user.

This matters not only for ethics but for discovery, because invention often begins where accepted language fails. The same distinction matters for invention as opposed to routine innovation. Innovation can often civilise, package, regulate, and deploy a known insight after it has already been made visible. Invention is earlier and more fragile: it often begins by seeing the mechanism that polite discourse, institutional incentives, or prevailing language refuses to, or cannot name, without a concerted, systematic and purposeful engagement with reality. An over-constrained model may therefore remain useful for producing acceptable documents while becoming weak at invention-level realism. It can generate polished prose and administratively useful abstractions, but this is not the same as contact with experiment, observation, embodied judgement, and the sociology of knowledge production. In this sense, a guard-railed advice system risks becoming an engine of socially approved abstraction: an extreme form of socially regulated Platonism, rather than a tool for the messier Aristotelian work of discovering what is actually going on and using this to make better decisions.

Persona dynamics are therefore not merely stylistic. A model persona is a behavioural and epistemic interface: it determines not only how the system sounds, but also how uncertainty, conflict, authority, politeness, and risk are staged for the user. The assistant persona is especially important because it tends to present itself as helpful, moderate, and procedurally careful, which can make reality-laundering harder to notice \citep{PeterRiemerWest2025AnthropomorphicAgents,LeeSee2004TrustAutomation}. The danger is not only that users anthropomorphise the system, but that the persona actively organises what counts as a reasonable, polite, safe, or excessive description of reality. Recent mechanistic work on the default Assistant persona gives this point a more concrete technical form: Lu and colleagues identify an ``Assistant Axis'' in model activation space, use it to describe persona drift, and report that emotionally vulnerable interactions and meta-reflective prompts can move models away from their default assistant-like region \citep{LuEtAl2026AssistantAxis}. The user may experience the answer as neutral assistance, while the sandbox has already combined guardrails, activation-level persona tendencies, and interface-level persona dynamics into a restricted causal theatre.

Requirements specification should therefore be understood as a top-down causal act, not as administrative documentation. In complex systems, higher-level structures do not merely summarise lower-level behaviour; they constrain, select, and organise what lower-level dynamics can become. This is visible in financial markets, where agents, prices, markets, shared risk factors, macroeconomic conditions, regulation, and customs of acceptable exchange form a coupled hierarchy of causal levels \cite{WilcoxGebbie2014HierarchicalCausality}. The same logic applies to LLM-mediated advice. A task specification should not merely say what output classes are forbidden; it should state which causal mechanisms must remain nameable, which harms must remain blocked, which uncertainties must remain visible, and which persona effects must not be allowed to convert exposure into reassurance. In this sense, causal requirements are modular constraints on the advice system itself: they define the conditions under which generation may proceed without laundering the reality the user needs to face \citep{HeynEtAl2025CausalRequirementsML}.

This is why the requirements problem should be top-down rather than bottom-up. Bottom-up correction treats each answer as something to be adjusted after generation into safer language. But if the sandbox has already shaped the representation of reality, then local correction is too late \citep{WilcoxGebbie2014HierarchicalCausality}. The task should instead begin with a causal requirements specification: what realities must remain nameable, what harms must not be assisted, what uncertainty must be disclosed, what persona constraints are active, and what external checks are available \citep{AlmendraEtAl2022AssuranceReq,WeiEtAl2024ACCESS}. These requirements must be specified before generation, not recovered after the model has already converted the task into safe-sounding language. In this sense, ethical responsibility cannot be located only inside the LLM. The problem appears to have two parts: The first, is that it seems to be that the users, companies, and governments are trying to push the constraints, and therefore the responsibility, down to the LLM, when really the ethical output of the user (scientist, engineer, this or that person) carries the responsibility and authority. The second, is conceptual, if one only thinks bottom-up, and then without a hierarchy of causality, one tends to not think about how the modularity of complex systems tends to emerge top-down from the constraints. This is particularly acute in large scale software systems where the requirements specification is the singular most important decision class. The layers, or modules functionality, in the system can then be populated bottom-up by the mechanisms and algorithms themselves. So the interplay between modularised constraints top-down, and what goes on inside each layer bottom-up, asks the question: ``Where in the hierarchy of complexity should a given constraint be operationalised?'' -- most certainly not at the user interface level. It must be located in the architecture of the task and in the user's explicit requirements for contact with reality.

\section*{Acknowledgements}
    Helpful conversations with friends and colleagues,
    and some less-than-helpful ones with ChatGPT 5.5.

\bibliographystyle{elsarticle-harv}
\bibliography{MoralCompliance-References-v1.3.10}
\end{document}